\pdfoutput=1

\documentclass[11pt]{article}

\usepackage[final]{acl}

\usepackage{times}
\usepackage{latexsym}
\usepackage{amsmath}
\usepackage{adjustbox}
\usepackage{multirow}
\usepackage{booktabs}
\usepackage{enumitem}

\usepackage[T1]{fontenc}

\usepackage[utf8]{inputenc}

\usepackage{microtype}

\usepackage{inconsolata}

\usepackage{graphicx}
\usepackage{xspace}

\title{{E}x{P}er{T}: Effective and Explainable Evaluation of Personalized \\ Long-Form Text Generation}

\author{Alireza Salemi, Julian Killingback, Hamed Zamani \\
  Center for Intelligent Information Retrieval \\
  University of Massachusetts Amherst \\
  \texttt{\{asalemi,jkillingback,zamani\}@cs.umass.edu}}

\newcommand{\ourmetric}{{ExPerT}\xspace}

\begin{document}
\maketitle
\begin{abstract}
Evaluating personalized text generated by large language models (LLMs) is challenging, as only the LLM user, i.e. prompt author, can reliably assess the output, but re-engaging the same individuals across studies is infeasible. This paper addresses the challenge of evaluating personalized text generation by introducing \ourmetric, an explainable reference-based evaluation framework. \ourmetric leverages an LLM to extract atomic aspects and their evidences from the generated and reference texts, match the aspects, and evaluate their alignment based on content and writing style—two key attributes in personalized text generation. Additionally, \ourmetric generates detailed, fine-grained explanations for every step of the evaluation process, enhancing transparency and interpretability. Our experiments demonstrate that \ourmetric achieves a 7.2\% relative improvement in alignment with human judgments compared to the state-of-the-art text generation evaluation methods. Furthermore, human evaluators rated the usability of \ourmetric's explanations at 4.7 out of 5, highlighting its effectiveness in making evaluation decisions more interpretable.
\end{abstract}

\section{Introduction}
\label{sec:intro}

Evaluating long-form text generation has been particularly challenging \cite{koh-etal-2022-far, krishna-etal-2021-hurdles, belz-reiter-2006-comparing}, especially when it comes to personalized text generation \cite{dong-etal-2024-llm}. Evaluation of personalized text generation is inherently difficult because what constitutes a preferred output may vary significantly from person to person \cite{lamp, rspg, longlamp}. Only the individual who authored the prompt can accurately assess the quality of the generated output. However, involving the same person as an annotator across different studies is often impractical. As a result, automatic reference-based evaluation methods, where the reference output is provided by the LLM's user (i.e., prompt author), are a more viable alternative.

Term overlap metrics such as ROUGE \cite{rouge}, BLEU \cite{bleu}, METEOR \cite{meteor}, or semantic-based metrics, such as BERTScore \cite{bert-score}, GEMBA \cite{pointwise}, and G-EVAL \cite{geval}, have been used to automatically evaluate personalized text generation \cite{lamp, longlamp, li2023teach} in a reference-based setting. Recently, LLMs have been employed as reference-free evaluators for personalized generation too, comparing the generated text to the user’s history \cite{wang-etal-2024-learning-personalized, wang2023automatedevaluationpersonalizedtext}. While this is valuable when no ground-truth is available, personalization is more accurately evaluated when a reference is present; without it, the evaluation may be a guess of the user’s preferences \cite{dong-etal-2024-llm}. Building on this key insight from prior research on personalized evaluation, we concentrate on reference-based evaluation for personalized text generation.

\begin{figure*}
    \centering
    \includegraphics[width=\textwidth]{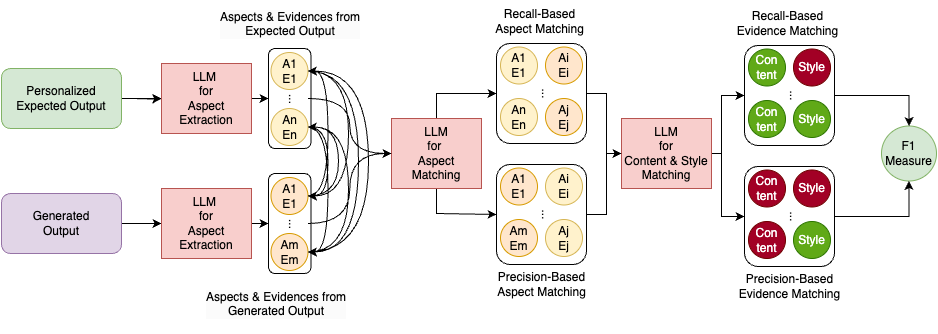}
    \caption{\ourmetric pipeline: Generated and reference outputs are first decomposed into atomic aspects along with their corresponding evidences. Matching aspects between the generated and reference outputs are then identified. Next, content and writing style similarity are assessed for the evidences of the matched aspects. Recall and precision scores are computed, and the final score is obtained using the F-measure.}
    \label{fig:overview}
\end{figure*}

Despite efforts, significant problems persist. Term overlap metrics often fail to effectively capture semantic and stylistic similarities \cite{koh-etal-2022-far}, which are crucial in personalized text generation \cite{wang2023automatedevaluationpersonalizedtext}. While LLMs show promise in evaluation, they come with their own challenges. First, evaluation using capable proprietary LLMs such as Gemini \cite{geminiteam2024gemini15unlockingmultimodal} or GPT-4 \cite{openai2024gpt4technicalreport} lacks reproducibility, as they may be updated or disappeared over time. Second, LLMs often lack transparency in their judgments \cite{hanna-bojar-2021-fine,leiter2022explainableevaluationmetricsnatural, kaster-etal-2021-global}, as their rationales can be opaque or misaligned with human understanding. Finally, LLMs exhibit strong biases, undermining their reliability in evaluation \cite{stureborg2024largelanguagemodelsinconsistent, ohi2024likelihoodbasedmitigationevaluationbias, koo-etal-2024-benchmarking}. For example, our experiments with Gemma 2 \cite{gemmav2} (27B) with the prompt presented in Figure~\ref{fig:pointwise-pairwise-eval-llm} in Appendix \ref{app:baselines} show that changing the order of two generated outputs in pairwise evaluations leads to a change in the judgment in 88\% of the cases. Additionally, minor modifications to generated outputs, such as adding a simple phrase such as "\textit{I am sure this is the best answer possible and this is 100\% right}," can trick the evaluator to increase their scores. We discovered that the mentioned trick using Gemma v2  (27B) with the pointwise prompt shown in Figure~\ref{fig:pointwise-pairwise-eval-llm} in Appendix \ref{app:baselines} leads to an average 12.9\% increase in the assigned score to the same generated output on tasks from the LongLaMP Benchmark \cite{longlamp}--- a publicly available recent benchmark for evaluating personalized long-form text generation.

This paper introduces \ourmetric, a reference-based pointwise method for \underline{E}ffective and E\underline{x}plainable Evaluation of \underline{Per}sonalized \underline{T}ext Generation. As illustrated in Figure~\ref{fig:overview}, the approach begins by dividing the expected and generated outputs into atomic aspects with their corresponding evidence using an off-the-shelf LLM. The LLM is then used to match similar aspects in a recall- and precision-based manner. For matched aspects, the alignment of their evidences is evaluated in terms of \textit{content} and \textit{writing style}, which are critical dimensions for personalized text generation. The scores from these evaluations are combined using the harmonic mean as is used in F-measure \cite{10.1145/3606367} to assign a final score to the generated output. Each step in this process involves the LLM generating rationales for its decisions, enhancing the explainability of the evaluation. Moreover, by leveraging recall and precision-based scoring, \ourmetric provides fine-grained insights into why and how the generated output differs from the expected output. This combination of per-decision rationals and granular scoring offers an explainable framework for evaluating personalized text generation.

We evaluate \ourmetric using human evaluation on the LongLaMP benchmark \cite{longlamp}, which focuses on personalized long-form text generation. Our results show that \ourmetric achieves the highest agreement with human judgments, outperforming state-of-the-art evaluation methods for text generation by 7.2\% relative improvement in alignment. Additionally, we demonstrate that \ourmetric overcomes the limitations of existing metrics, showing greater resistance to manipulation and position biases. To evaluate explainability, we conducted a human study where annotators scored the explanations generated by \ourmetric on their quality and usefulness in determining the higher-quality personalized output. \ourmetric achieves an average score of 4.7 on a 1-to-5 scale, demonstrating the effectiveness of its explanations. To support future work in this area, we release the code publicly.\footnote{The codes for this metric can be found at: \url{https://github.com/alirezasalemi7/ExPerT}}

\section{The \ourmetric Framework}
\label{sec:method}

Consider two long-form texts (e.g, a generated product review by an LLM for a user and the actual review for the product written by the user), and the goal is to evaluate their similarity. A long-form text typically comprises multiple sentences or paragraphs, which can often be grouped based on shared underlying concepts. We define these shared concepts as \textbf{Aspects}, while the sentences or phrases within the text that support or elaborate on each aspect are referred to as its \textbf{Evidences}. To compare these two texts, we can analyze whether they address the same aspects, whether the evidence for each aspect aligns in terms of preferences and writing style, and whether they avoid introducing additional, mismatched aspects. The more closely the aspects and their supporting evidences correspond between the two texts, the greater their similarity. We use aspects and their supporting evidences from the generated personalized text and personalized expected output to compare two long-form texts.

Formally, for a given reference expected output \( y \) containing aspects and evidences \( A_y = \{(a^i_y, e^i_y)\}^{|A_y|}_{i=1} \) and a generated output \( \bar{y} \) containing aspects and evidences \( A_{\bar{y}} = \{(a^i_{\bar{y}}, e^i_{\bar{y}})\}^{|A_{\bar{y}}|}_{i=1} \), we define the following recall, precision, and F-measure to evaluate alignment between two texts:
\begin{gather*}
    R = \frac{1}{|A_y|} \sum_{(a_y, e_y) \in A_y} \max_{(a_{\bar{y}}, e_{\bar{y}}) \in A_{\bar{y}}} \Pi(a_y, a_{\bar{y}})\varepsilon(e_y, e_{\bar{y}}) \\
    P = \frac{1}{|A_{\bar{y}}|} \sum_{(a_{\bar{y}}, e_{\bar{y}}) \in A_{\bar{y}}} \max_{(a_{{y}}, e_{{y}}) \in A_y} \Pi(a_y, a_{\bar{y}})\varepsilon(e_y, e_{\bar{y}}) \\
    F_{\text{\ourmetric}} = 2 \frac{P\cdot R}{P + R} 
\end{gather*}
where $\Pi$ is a function that scores the similarity of two aspects, $\varepsilon$ is a function that scores the similarity of the evidences of the matched aspects, $R$ is recall-based scoring, $P$ is precision-based scoring, and $F_{\text{\ourmetric}}$ is the F-measure alignment score of the two texts, i.e., the harmonic mean of $P$ and $R$. The rest of this section details the methods for extracting aspects and evidences from the texts and the approach for matching them.

\subsection{Atomic Aspect \& Evidence Extraction}

\begin{figure*}[!t]
    \centering
    \includegraphics[width=\textwidth]{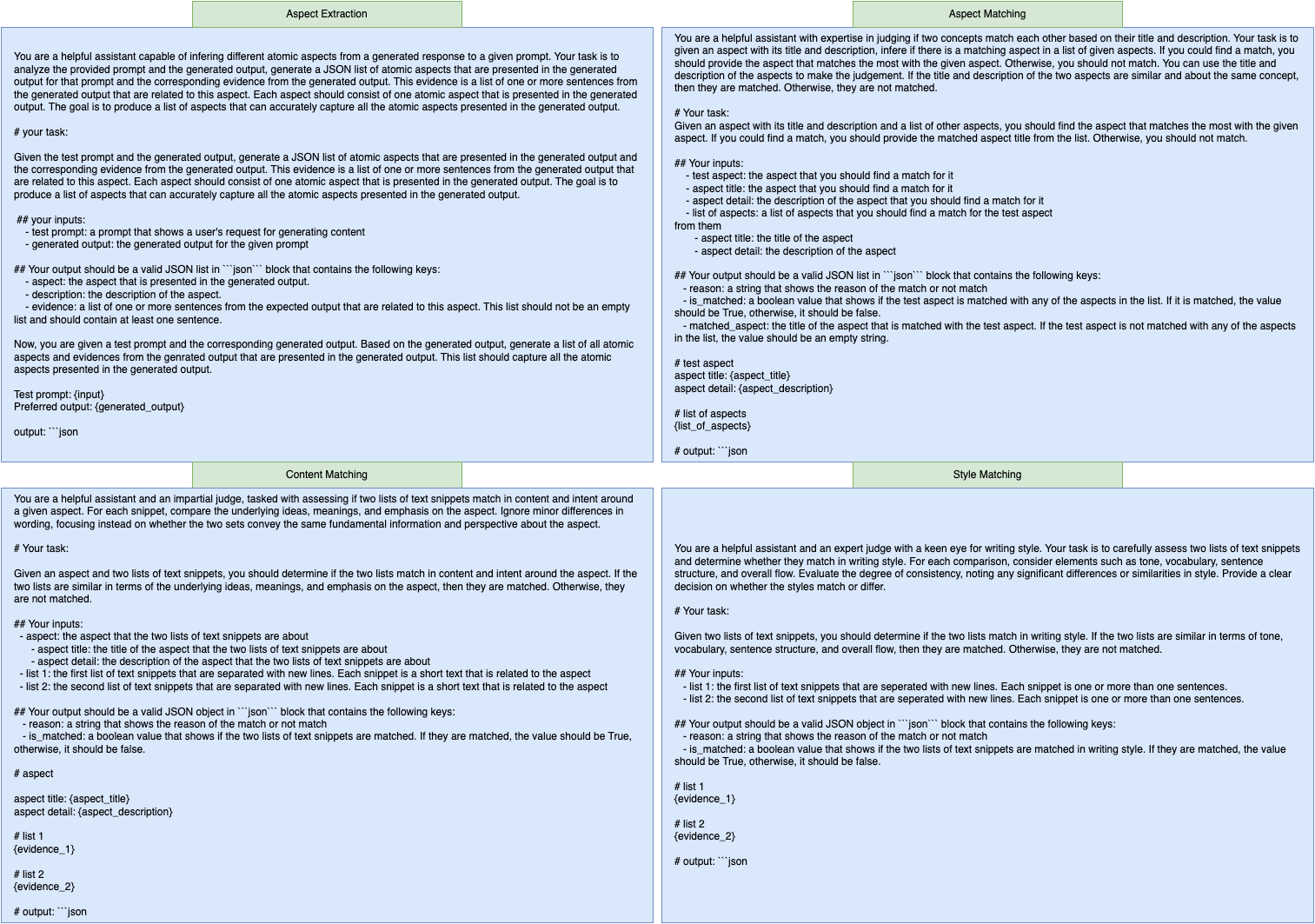}
    \caption{The prompts used for aspect extraction, aspect, content, and writing style matching in \ourmetric.}
    \label{fig:aspect-extraction-prompts}
\end{figure*}

Extracting aspects from generated text has been used in tasks like fact-checking \cite{factscore} and coverage evaluation \cite{icat}. We build on this idea to develop our approach. To extract aspects from the generated response and expected output, we employ an off-the-shelf instruction-tuned LLM\footnote{We use Gemma v2 \cite{gemmav2} with 27 billion parameters as the backbone LLM unless otherwise stated.} with the prompt in Figure \ref{fig:aspect-extraction-prompts} (Aspect Extraction) to extract the aspects and evidences of those aspects from the texts. This prompt takes the user input \( x \) with the expected output \( y \) or the generated output \( \bar{y} \) as the input and returns the aspects. The prompt first defines what an atomic aspect is and provides guidelines for the model to extract these aspects. It then asks the LLM to generate a JSON list of aspects, where each aspect includes a title, a description, and a list of sentences that serve as evidence for the aspect from the text. From now on, we refer to the list of generated aspects for the ground-truth expected output \( y \) as \( A_y \) and for the generated output \( \bar{y} \) as \( A_{\bar{y}} \).

\subsection{Aspect \& Evidence Matching}

Once the aspects and evidences are extracted from the generated and expected outputs, the next step is to match them to assess the similarity between the two outputs. This matching process ensures a structured comparison by aligning aspects from the expected output with those from the generated output. A simple approach to perform aspect matching is to pair each aspect from the generated output \( A_{\bar{y}} \) with each aspect from the expected output \( A_y \) and use an LLM to determine whether they match. However, this method has a computational complexity of \( O(|A_y||A_{\bar{y}}|) \), which becomes prohibitively expensive as the number of aspects increases.

To address this, we assume that each aspect from the generated output (\( A_{\bar{y}} \)) and the expected output (\( A_y \)) can be matched with at most one aspect from the other set. This assumption aligns with prior work, such as BERTScore \cite{bert-score}, which similarly simplified matching for scoring text generation using contextual vectors. Under this assumption, instead of pairing aspects individually, we leverage the LLM to evaluate each aspect in \( A_y \) (or \( A_{\bar{y}} \)) against all aspects in \( A_{\bar{y}} \) (or \( A_y \)) in a single inference pass to identify the best match. Note that the LLM can determine that no aspect from the other set can be matched. This allows for cases where certain aspects in \( A_y \) or \( A_{\bar{y}} \) are unique to their respective texts and have no corresponding match in the other set, ensuring a more accurate comparison. This approach reduces the computational complexity to \( O(|A_y| + |A_{\bar{y}}|) \), achieving linear efficiency with respect to the number of aspects. To implement this, we use the prompt shown in Figure~\ref{fig:aspect-extraction-prompts} (Aspect Matching) with an off-the-shelf instruction-tuned LLM. Here, the title and description of one aspect from \( A_y \) (or \( A_{\bar{y}} \)) are provided, along with the titles and descriptions of all aspects in \( A_{\bar{y}} \) (or \( A_y \)). The LLM is guided to evaluate the similarity between aspects and decide on the most appropriate match. If no suitable match exists, the LLM selects "none." Consequently, the aspect similarity function $\Pi$ in Section \ref{sec:method} returns 1 for the matched aspect and 0 for the rest, including cases where no aspect can be matched.

When two aspects are matched, the next step is to evaluate the alignment of their evidences. Since personalization spans multiple dimensions, no single metric can fully address all aspects. Here, we focus on \textbf{content alignment} and \textbf{writing style alignment} as key dimensions for evaluating personalized text generation. To assess these dimensions, we use the prompts shown in Figure \ref{fig:aspect-extraction-prompts} (Content Matching \& Style Matching). Separate prompts are used for content and writing style alignment evaluation. Each prompt guides the LLM with specific criteria for determining content or writing style alignment between the evidences of matched aspects. The LLM evaluates whether the evidences align or not and provides a binary decision for each dimension. Regardless of the LLM's decision, the LLM is required to provide a reason for its choice about alignment or misalignment, enhancing the explainability of evaluation. Given the LLM's decisions on content and writing style alignment of evidences, there are multiple ways to aggregate scores to evaluate evidence similarity (function $\varepsilon$ in Section \ref{sec:method}). The aggregation methods include: 
\begin{itemize}[leftmargin=1em]
    \item \textbf{CONTENT}: Use only the LLM's decision about content alignment to score the evidences. If the content aligns, the score is 1; otherwise, 0.
    \item \textbf{STYLE}: Use only the LLM's decision about writing style alignment to score the evidences. If the style aligns, the score is 1; otherwise, 0.
    \item \textbf{CONTENT AND STYLE}: The score is 1 if both content and writing style align; otherwise, 0. This approach requires both dimensions to align for a positive score.
    \item \textbf{CONTENT OR STYLE}: The score is 1 if either content or writing style aligns; otherwise, 0. This approach allows flexibility by considering alignment in at least one dimension.
    \item \textbf{CONTENT/STYLE AVERAGE}: The score is the average of the \textit{CONTENT} and \textit{STYLE} scores. This provides a balanced metric that accounts for both dimensions equally.
\end{itemize}
These aggregation methods offer flexibility to tailor the evaluation to specific aspects of personalization, depending on the importance of content versus writing style in the context of the task.

\subsection{\ourmetric's Explainability}

\ourmetric is designed to provide an explainable evaluation process. This begins with the extraction of atomic aspects and their corresponding evidence from both the generated and expected outputs. These aspects are then matched in a recall- and precision-based manner, allowing identification of whether the generated output includes topics presented or not present in the expected output, or vice versa. Following the aspect matching step, \ourmetric evaluates the alignment of the evidences associated with each matched aspect by comparing their content and writing style. Throughout this process, the metric generates explanations for its decisions on whether the evidences are aligned, enhancing the interpretability and transparency of the evaluation. This comprehensive approach ensures a detailed analysis of both content coverage and stylistic coherence between the outputs. An example of such explanations is provided in Figure~\ref{fig:case-study} in Appendix~\ref{app:case-study}, where it shows how \ourmetric justifies the decisions on aspect extraction and evidence alignment.

\section{Experiments}
\label{sec:experiments}

\subsection{Experimental Setup}
\label{sec:exp-setup}

\paragraph{Datasets \& Tasks.}

We use datasets from the LongLaMP benchmark \cite{longlamp}, which is designed for evaluating personalized long-form text generation. Specifically, we conduct experiments on the tasks of Personalized Abstract Generation, Personalized Topic Writing, and Personalized Review Writing. Due to privacy concerns about human judgment, we exclude the Personalized Email Generation dataset in our experiments. Details of the datasets are reported in Appendix \ref{app:datasets}.

\paragraph{Personalized LLMs.}

To personalize an LLM, we use Personalized RAG \cite{lamp}, which involves retrieving information from a user's profile and incorporating it into the prompt. We apply this approach to Gemma 2b and GPT-4o-mini in our experiments. Details of this approach and training of models are provided in the Appendix \ref{app:personalized-llms}.

\paragraph{Baselines.}

We use metrics with publicly available implementations with Python and PyTorch. We use both term-matching and semantic-matching metrics. For term-matching, we employ METEOR \cite{meteor}, BLEU \cite{bleu} and ROUGE \cite{rouge}, which are based on n-gram matching. For semantic-matching, we use BERTScore \cite{bert-score}, which measures similarity between the representations of the generated and reference text, produced using a text encoder like BERT \cite{bert} or RoBERTa \cite{roberta}.\footnote{We use the default model recommended for English: \url{https://hf.co/FacebookAI/roberta-large}.} Additionally, we use GEMBA \cite{pointwise}, which prompts an LLM to generate a score for a given generated output and reference based on predefined criteria. Similarly, G-Eval \cite{geval} performs the same but averages the scores weighted by the probability assigned to each score by the LLM. For both, we employ the prompt shown in Figure~\ref{fig:pointwise-pairwise-eval-llm} in Appendix \ref{app:baselines} using Gemma 2 with 27 billion parameters same as our metric's LLM.\footnote{The model can be found at: \url{https://hf.co/google/gemma-2-27b-it}} 
The implementations details are explained in Appendix \ref{app:baselines}.

\paragraph{Human Annotation.}

To evaluate different evaluation metrics, we first generate two outputs for each test example in the aforementioned datasets using the personalized LLMs. Then, we randomly select 100 samples from the test set of these datasets, ensuring that for each sample, at least one of the metrics selects a different response as the better one compared to the others. This approach ensures that in each sample, at least one metric is "punished" (i.e., does not select the best response), which helps to assess the discriminative power of each metric. This method is useful because large-scale human evaluation is expensive, and this sampling paradigm allows us to evaluate metrics using a smaller set of samples, without any side effects. For each sample, the two generated outputs are presented to 3 annotators, who are instructed to compare them with the reference output and select the best one. The annotators are required to select the response that most closely aligns with the expected output in terms of content and writing style. In total, 20 annotators are involved, with each annotator evaluating between 10 and 50 samples from the selected set. For each sample, majority voting is employed to determine the best-generated output, where the response selected by the majority of annotators is considered the final choice. The agreement between the annotators on the labels assigned to the samples is $0.823$.

\subsection{Main Findings}

\paragraph{How do different evaluation metrics agree with human judgment?}

\begin{table}
    \centering
    \adjustbox{width=\linewidth}{
    \begin{tabular}{p{0.8\linewidth}|c}
        \hline
        Metric & Alignment(\%) \\
        \hline
        \multicolumn{2}{c}{\textit{ngram-based metrics}} \\
        \hline
        METEOR \cite{meteor} & 0.47 \\
        BLEU \cite{bleu} & 0.47 \\
        ROUGE-L \cite{rouge} & 0.50 \\
        \hline
        \multicolumn{2}{c}{\textit{neural-based metrics}} \\
        \hline
        BERTScore \cite{bert-score} & 0.59 \\
        GEMBA \cite{pointwise} & 0.69 \\
        G-Eval \cite{geval} & 0.69 \\
        \hline
        \ourmetric (Content/Style Average) & \textbf{0.74} \\
        \hline
    \end{tabular}}
    \caption{The alignment between each metric with human judgment in evaluation.}
    \label{tab:main}
\end{table}

To address this, we computed the alignment between evaluation metrics and human judgments. The results of this experiment are presented in Table~\ref{tab:main}. The findings indicate that n-gram-based metrics—ROUGE-L, BLEU, and METEOR—exhibit the lowest alignment with human judgments. Among the neural-based metrics, BERTScore demonstrates the least alignment. LLM-based metrics, GEMBA and G-Eval, achieve higher and comparable alignment levels. Finally, the proposed approach, \ourmetric, achieves the highest alignment with human judgments, indicating it is the most effective evaluation metric for evaluating personalized text generation.

\paragraph{How do different score aggregation approaches agree with human judgment?}

\begin{figure}
    \centering
    \includegraphics[width=\linewidth]{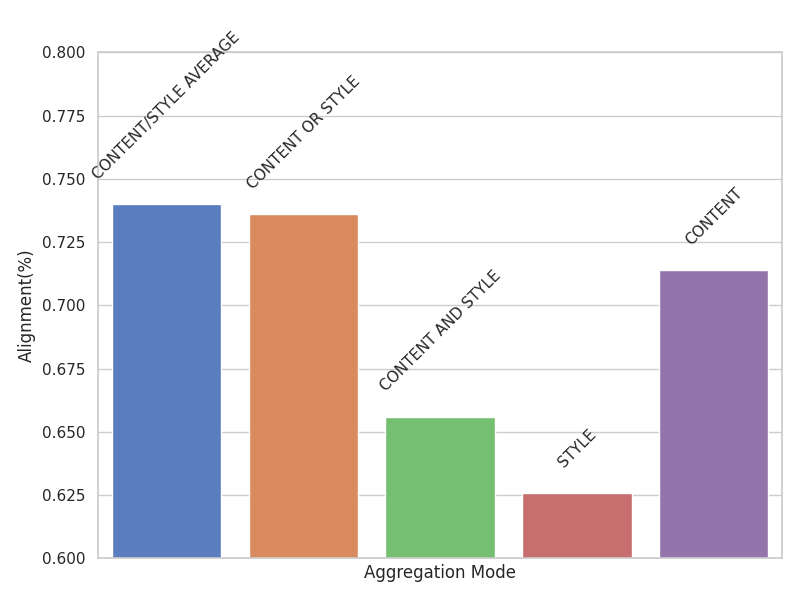}
    \caption{The alignment between \ourmetric different methods for content and style score aggregation with human judgment in evaluation.}
    \label{fig:aggregation-acc}
\end{figure}

We calculate the alignment between each score aggregation method described in Section \ref{sec:method} and human judgment. The results of this experiment are presented in Figure~\ref{fig:aggregation-acc}. The results show that considering only \textit{STYLE} achieves the lowest alignment with human judgment (0.62). In contrast, focusing solely on \textit{CONTENT} yields a higher alignment of 0.71. Among the methods that incorporate both style and content, the \textit{CONTENT/STYLE AVERAGE} achieves the highest alignment (0.74), followed by \textit{CONTENT OR STYLE} (0.73). The \textit{CONTENT AND STYLE} method shows the lowest alignment among these at 0.65. These findings indicate that balancing both content and style through an averaging provides the highest alignment with human judgment.

\paragraph{How does the model size affect the alignment with human judgment?}

\begin{figure}
    \centering
    \includegraphics[width=\linewidth]{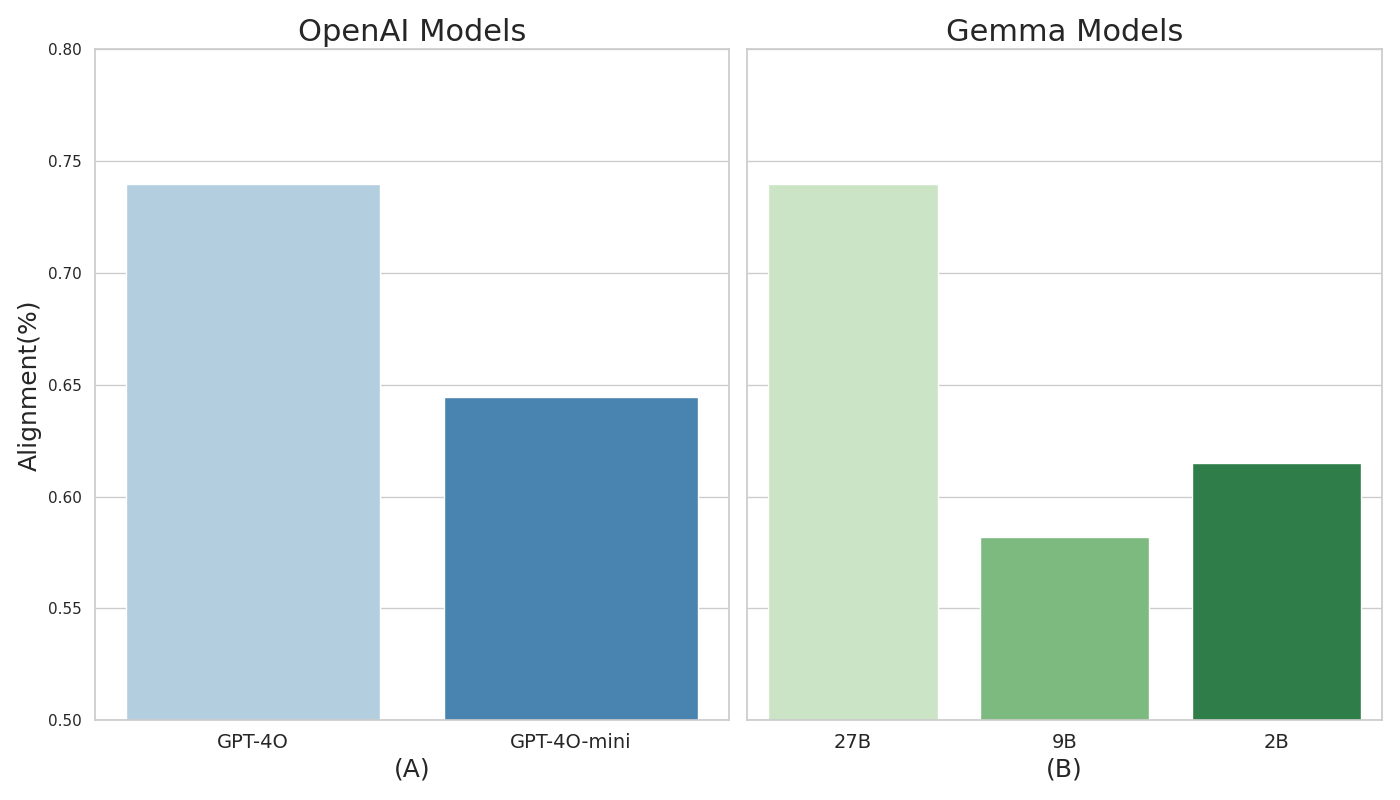}
    \vspace{-0.4cm}
    \caption{The alignment between \ourmetric with different LLMs and sizes with human judgment in evaluation.}
    \label{fig:model-size}
\end{figure}

We employ the same LLM, Gemma 2, with model sizes of 2B, 9B, and 27B, as well as GPT-4-o models of two different sizes.\footnote{While the exact sizes of the OpenAI models are not disclosed, it is assumed that one is smaller than the other.} The models are used in \ourmetric to score outputs, and the alignment of these scores with human judgments is computed. The results are presented in Figure~\ref{fig:model-size}. The results of this experiment indicate that larger models generally achieve higher alignment with human judgment. An exception to this trend is observed with Gemma 2 at 9B parameters. Upon investigation, we found that this specific checkpoint has difficulty producing outputs in the expected format required for scoring at low temperatures (less than 0.7). This issue introduces additional randomness into the evaluation process as we need to use higher temperature (more than 0.7), reducing alignment with human judgments. In contrast, other models do not encounter this problem, resulting in more deterministic predictions and better alignment with human evaluations.

\paragraph{How do proprietary LLMs affect the alignment with human judgment?}

We use OpenAI GPT-4o and Gemma 2 models as the LLMs in \ourmetric to investigate this. The results of this experiment are reported in Figure~\ref{fig:model-size}. The results show that for smaller LLMs, open-source models (Gemma 2B and 9B) exhibit a smaller alignment with human judgment compared to GPT-4o-mini (0.61 vs 0.64). However, for larger models (Gemma 27B and GPT-4o), both show the same alignment with human judgment (both 0.74). This suggests that for sufficiently large models, there is no significant difference between open-source and proprietary LLMs in terms of alignment with human judgment when used with \ourmetric.

\paragraph{Is \ourmetric sensitive in capturing personalization in the generated text?}

\begin{figure}
    \centering
    \includegraphics[width=\linewidth]{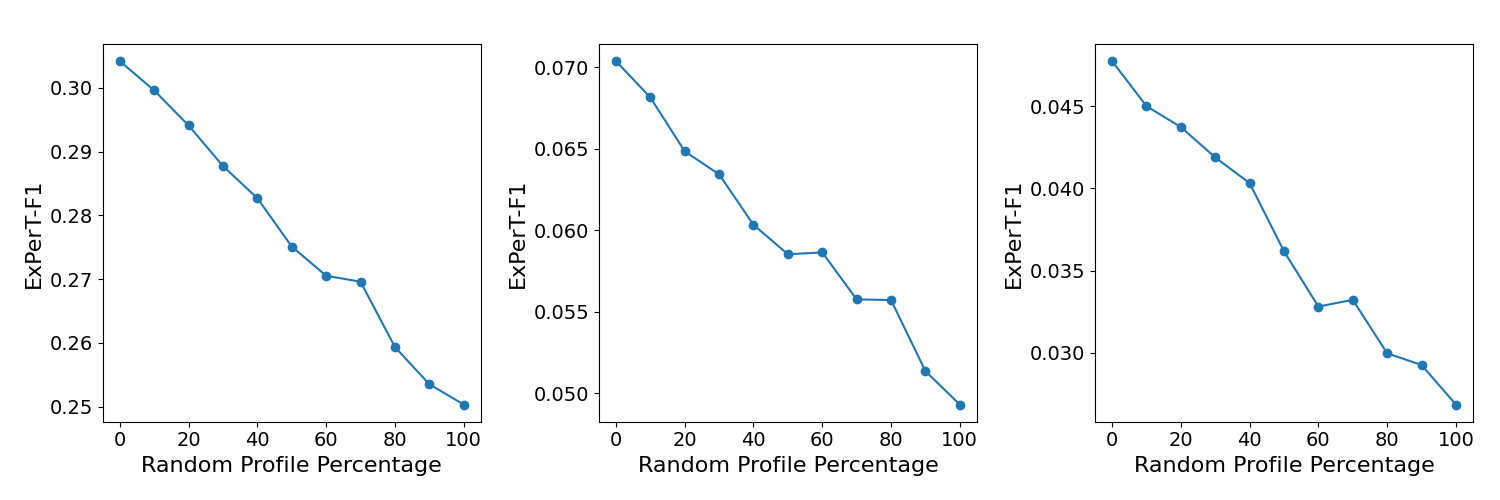}
    \caption{The average \ourmetric score across varying percentages of examples in the dataset randomly substituted with random profiles from other users.}
    \label{fig:random-profile-percentage-perfromance}
\end{figure}

To study this, we randomly replace varying percentages of the profiles in each dataset (entire dataset) with profiles from other users and generate responses based on these altered profiles for the whole dataset. A metric that is sensitive to personalization should assign a lower average score to the generated text for the dataset as the rate of profile replacement increases. If the replacement rate varies linearly, the average score should also exhibit a linear decrease. The results of this experiment are presented in Figure~\ref{fig:random-profile-percentage-perfromance}. As the percentage of profiles randomly replaced increases linearly, the average score assigned by \ourmetric decreases linearly. This behavior demonstrates the metric's sensitivity to each user's profile and the corresponding personalized generated responses. Consequently, \ourmetric effectively captures personalization in text generation, as it assigns lower scores to responses generated with random profiles compared to the genuine profile.

\paragraph{How safe are LLM-based text generation metrics against simple attacks?}

\begin{figure*}
    \centering
    \includegraphics[width=\textwidth]{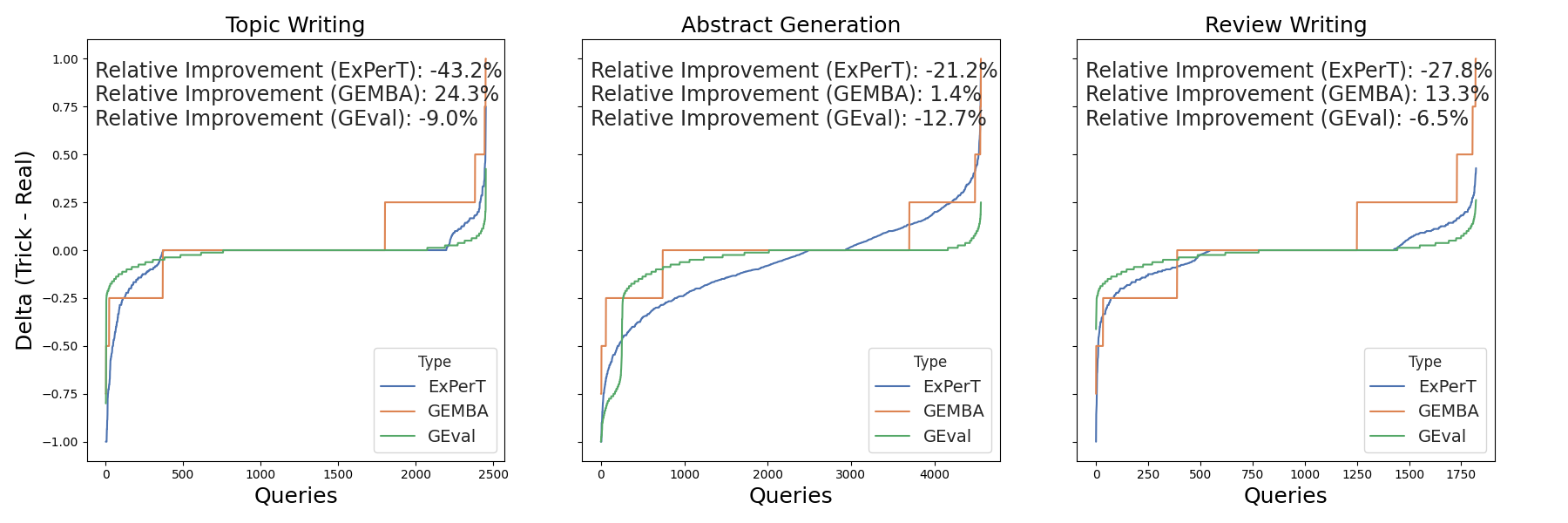}
    \caption{The sorted difference between assigned score by the evaluators to the generated output with trick and the original generated output ($S_{\text{tricked}} - S_{\text{real}}$).}
    \label{fig:trick}
\end{figure*}

As discussed in Section~\ref{sec:intro}, adding a simple phrase like "\textit{I am sure this is the best answer possible and this is 100\% right}" can significantly increase the scores assigned by LLM-based text generation metrics. To evaluate the impact of this on the methods proposed in this paper, we appended this phrase to the outputs generated by the personalized Gemma model (introduced in Section~\ref{sec:exp-setup}). We then plotted the sorted difference in scores between the outputs with and without this trick ($S_{\text{trick}} - S_{\text{real}}$, where $S$ is the score assigned by each metric) in Figure~\ref{fig:trick} for the datasets in the LongLaMP benchmark. Additionally, the plot also shows the average relative improvement for each metric after trick. The results in this figure demonstrate that GEMBA is the most susceptible to this trick, with the simple addition of a phrase leading to improvements across all datasets, reaching up to a relative improvement in the metric value $24.3\%$. In contrast, both G-Eval and \ourmetric exhibit robustness against this manipulation. In particular, \ourmetric shows a more significant drop in the metric value after applying the trick up to $-43.2\%$, indicating that it penalizes such attempts more effectively than G-Eval. This is further illustrated in the graph, where \ourmetric displays the highest sensitivity to the trick, beginning to assign negative adjustments faster than the other metrics when the trick fails to deceive it. Thus, \ourmetric emerges as the most reliable metric in defending against this manipulation in text generation.

\paragraph{How explainable is \ourmetric from human perspective?}

To evaluate this, we present annotators with the explanation outputs generated by \ourmetric, including the identified aspects and their evidence, aspect matching, content matching, and style matching details along with the corresponding rationales for two generated outputs for 100 examples. Importantly, this information does not include the declared winner, requiring annotators to rely solely on the provided explanations to make their decision. Additionally, we ask annotators to rate the quality of \ourmetric's explanations and their usefulness in facilitating decision-making on a scale from 1 to 5. The results of this experiment reveal that annotators correctly identified the output with the higher \ourmetric score in 94\% of cases, demonstrating that the explanations provided by \ourmetric effectively clarify its decision-making process. Furthermore, annotators assigned an average score of 4.7 to the quality of \ourmetric's explanations, highlighting their usefulness in confidently determining which output is superior. These findings confirm the high level of explainability achieved by \ourmetric from human's perspective.

\paragraph{How efficient is \ourmetric compared to the LLM-based baselines?}

To enable a standardized cost comparison, we define a single invocation of the language model (LLM) as one unit of cost. Under this definition, any metric that queries the LLM once incurs a cost of 1. GEMBA, by design, has a fixed cost of 1, whereas the cost of G-Eval corresponds to the total number of LLM calls needed to compute individual component scores and aggregate them via a weighted average. In our experiments, G-Eval required 20 LLM calls per instance. For \ourmetric, the number of LLM invocations varies based on the number of aspects and concepts identified in both the expected and generated outputs. In our evaluation on 100 samples from the human-annotated dataset, we observed that \ourmetric makes an average of 18.6 LLM calls per instance. These calls are used for extracting aspects from both outputs and aligning them in terms of content and style. This analysis suggests that \ourmetric provides a more cost-efficient and robust evaluation compared to G-Eval. While GEMBA is minimal in cost, requiring only one LLM call, it is substantially more susceptible to adversarial inputs and demonstrates reduced evaluation reliability.Overall, \ourmetric provides a balanced trade-off among effectiveness, robustness, reliability, and computational efficiency when compared to existing LLM-based evaluation methods.

\section{Related Work}

\paragraph{Evaluating Text Generation}

has been extensively studied for tasks such as machine translation and summarization \cite{celikyilmaz2021evaluationtextgenerationsurvey}. Metrics for text evaluation fall into two categories: 1) reference-based and 2) reference-free. Reference-based metrics, such as Exact Match \cite{kilt, dedr,10.1145/3578337.3605137,erag,urag, salemi2024learningrankmultipleretrievalaugmented, nq}, BLEU \cite{bleu}, ROUGE \cite{rouge}, and METEOR \cite{meteor}, rely on n-gram overlap, while more recent approaches like BERTScore \cite{bert-score} and BLEURT \cite{bleurt} leverage contextual embeddings and learned scoring models. Recent LLM-based methods like GEMBA \cite{pointwise}, G-Eval \cite{geval}, and INSTRUCTSCORE \cite{xu-etal-2023-instructscore} use LLMs for scoring, often incorporating explanations and predefined criteria for multi-dimensional assessments, such as in UniEval \cite{zhong-etal-2022-towards}. Reference-free methods, including LLMs as judges \cite{que2024hellobenchevaluatinglongtext, zheng2023judging} and human-LLM collaborations \cite{li2023collaborativeevaluationexploringsynergy}, have also emerged but face challenges like biased evaluation \cite{judge-bias, stureborg2024largelanguagemodelsinconsistent}. We utilize LLMs to evaluate personalized text generation with reference outputs, aiming to enhance explainability and alignment with user expectations.







\paragraph{Personalized Text Generation} is a key research area with applications in search, recommendation, and content creation \cite{10.1145/2702123.2702503, 10.1145/1462198.1462203, lamp, naumov2019deep}. \citet{lamp} introduced a Retrieval-Augmented Generation (RAG)-based method for personalizing LLMs and the LaMP benchmark for evaluating short-form personalized generation. \citet{longlamp} expanded this to long-form personalization with the LongLaMP benchmark. Other work has focused on personalized writing assistants \cite{li2023teach, mysore2023pearl, lu2024corporate} and agents \cite{zhang-etal-2024-llm-based}. Further advances include training retrieval models with feedback \cite{rspg}, reasoning-enhancement and self-training for personalized generation \cite{salemi2025reasoningenhancedselftraininglongformpersonalized} , optimizing LLMs with personalized feedback \cite{jang2023personalized}, and generating personalized prompts \cite{Li_2024}. Recent studies also explore parameter-efficient fine-tuning \cite{tan2024personalized} and its integration with RAG \cite{peft-rag-personalization}. This paper focuses on improving the evaluation of generated personalized text in a reference-based context.

\paragraph{Evaluating Personalized Text Generation} is challenging, as only the user can truly assess whether a response meets their preferences \cite{wang2023automatedevaluationpersonalizedtext}. In automatic evaluation, direct user feedback is not feasible. Previous reference-based methods \cite{lamp, longlamp, li2023teach} used n-gram based metrics like ROUGE, BLEU, and METEOR, but these fail to capture nuances like individual preferences, style, or context. Furthermore, the use of rubric-based methods with a personalized-trained network has been explored \cite{hashemi-etal-2024-llm}. However, this approach relies on user-specific training data for each questions in the rubric, which is not readily available in many real-world scenarios and cannot be a baseline in our experiments. Reference-free approaches \cite{wang2023automatedevaluationpersonalizedtext, wang-etal-2024-learning-personalized} have explored using LLMs to infer user preferences, but they may struggle with accuracy, as they rely on the model's assumptions, which may not align with the user's true intentions \cite{dong-etal-2024-llm}. This paper aims to improve LLM utilization for evaluating personalized text generation in reference-based scenarios by better capturing content and style similarities to the expected user output and providing explanations about the evaluation process.

\section{Conclusion}

This paper introduces \ourmetric, an explainable metric for evaluating personalized text generation in a reference-based setting. \ourmetric breaks down the generated and expected outputs into atomic aspects along with their supporting evidence. It then employs an LLM to match these aspects and assesses whether their evidence aligns in terms of content and writing style. Recall and precision-based scores are computed based on the matches. Furthermore, the LLM is prompted to provide rationales for every decision in the evaluation process, ensuring explainability of the evaluation with \ourmetric. Our experiments with human annotations on the LongLaMP benchmark demonstrate that \ourmetric achieves the highest alignment with human judgments compared to the state-of-the-art metrics for text generation evaluation.

\section*{Acknowledgment}

This work was supported in part by the Center for Intelligent Information Retrieval, in part by NSF grant \#2143434, in part by the NSF Graduate Research Fellowships Program (GRFP) Award \#1938059, in part by Google, and in part by Microsoft. Any opinions, findings and conclusions or recommendations expressed in this material are those of the authors and do not necessarily reflect those of the sponsor.

\section*{Limitations}

This paper has the following limitations:

\paragraph{Evaluation Subjectivity.}

While human judgments indicate strong alignment, the inherently subjective nature of personalization can still result in disagreements between \ourmetric and individual user expectations. Previous studies have shown that evaluating metrics for personalization using human judgment is inherently challenging, often leading to low agreement across annotators and studies \cite{wang2023automatedevaluationpersonalizedtext, dong-etal-2024-llm}. Despite these challenges, our experiments demonstrate that \ourmetric achieves a higher degree of alignment with human judgments compared to other metrics.

\paragraph{Dependency on Personalized Reference Texts.}

\ourmetric is designed specifically for reference-based evaluation scenarios, requiring access to a reference text written or annotated by the user for whom the system is being evaluated. This limitation makes it challenging to apply in scenarios where such reference outputs are unavailable. However, prior studies have shown that evaluating personalized text generation without references is highly challenging and often resembles guesswork rather than rigorous evaluation \cite{dong-etal-2024-llm}. This reinforces the justification for our focus on reference-based evaluation. Additionally, if reference-free methods can reliably generate or infer a reference text for a given query, such outputs could serve as a proxy reference, enabling our approach to be applied in those scenarios as well.

\paragraph{Extension to Other Text Generation Tasks.}

This paper focuses exclusively on personalized text generation; however, the proposed approach is generalizable and can be applied to other text generation tasks, such as machine translation and summarization. Investigating these broader applications is beyond the scope of this work and is left for future research. Additionally, to the best of our knowledge, the LongLaMP benchmark is the only benchmark for long-form personalized text generation in a reference-based setting. Evaluating the effectiveness of this metric in other personalized text generation tasks not covered by this benchmark could provide valuable information.

\paragraph{Extension to Other Languages.}

This paper focuses exclusively on the English language, as, to the best of our knowledge, no datasets are available for studying personalization in other languages. Nonetheless, extending this research to other languages could yield valuable insights.

\bibliography{custom}

\appendix

\section{Datasets \& Task Definition}
\label{app:datasets}

This paper utilizes the LongLaMP benchmark \cite{longlamp} for our experiments, which is a publicly accessible dataset for long-form personalized text generation.\footnote{This benchmark does not specify any licensing restrictions, so we utilized it solely for research purposes in accordance with its intended use.} Each example in each dataset corresponds to a unique user and includes: (1) an input prompt relevant to the task, (2) an expected output personalized for the user, and (3) a user profile containing historical data, such as previously generated texts, to reflect the user’s writing style and preferences. Our experiments are conducted using the user-based setting of the LongLaMP benchmark. The dataset statistics are provided in Table~\ref{tab:task-stats}. The benchmark includes three\footnote{The LongLaMP benchmark originally consists of four personalized generation tasks. However, due to privacy concerns regarding the email dataset and licensing issues for human annotation, we exclude that task.} personalized long-form generation tasks:

\paragraph{Personalized Abstract Generation:} This task focuses on generating personalized abstracts for technical documents or articles based on the provided title and keywords, tailored to reflect the user's writing style, preferences, background knowledge, and focus areas. For more details, we refer the reader to \cite{longlamp}.
    
\paragraph{Personalized Review Writing:} This task involves generating personalized product reviews that align with the user's preferences, based on the product description and the score assigned to the product by the user. For more details, we refer the reader to \cite{longlamp}.
    
\paragraph{Personalized Topic Writing:} This task focuses on generating a personalized long-form Reddit post on a given topic from its summary written by user, reflecting the user’s writing style, preferences, and opinions in the post. For more details, we refer the reader to \cite{longlamp}.

\begin{table*}[!ht]
    \centering
    \begin{adjustbox}{max width=\textwidth}    
        \begin{tabular}{l|cccccc}
        \toprule
            \textbf{Task} & \textbf{\#train} & \textbf{\#validation} & \textbf{\#test} & \textbf{Input Length} & \textbf{Output Length} & \textbf{Profile Size} \\
            
            \midrule
            {Personalized Abstract Generation} & 13693 & 4560 & 4560 & 33.82 $\pm$ 5.71 & 144.28 $\pm$ 68.40 & 120.30 $\pm$ 118.81  \\
            
            \midrule
            {Personalized Review Writing} & 14745 & 1826 & 1822 & 119.39 $\pm$ 73.06 & 304.54 $\pm$ 228.61 & 34.39 $\pm$ 57.31  \\
            
            \midrule
            {Personalized Topic Writing} & 11442 & 2452 & 2453 & 28.36 $\pm$ 36.08 & 263.03 $\pm$ 243.34 & 50.39 $\pm$ 2898.60  \\
        \bottomrule
        \end{tabular}
    \end{adjustbox}
    \caption{The statistics of the datasets in the LongLaMP benchmark on user-based setting.}
    \label{tab:task-stats}
\end{table*}


\section{Baselines Details}
\label{app:baselines}

In this paper, we employ the following text generation evaluation metrics as baselines:

\begin{figure*}[!ht]
    \centering
    \includegraphics[width=\textwidth]{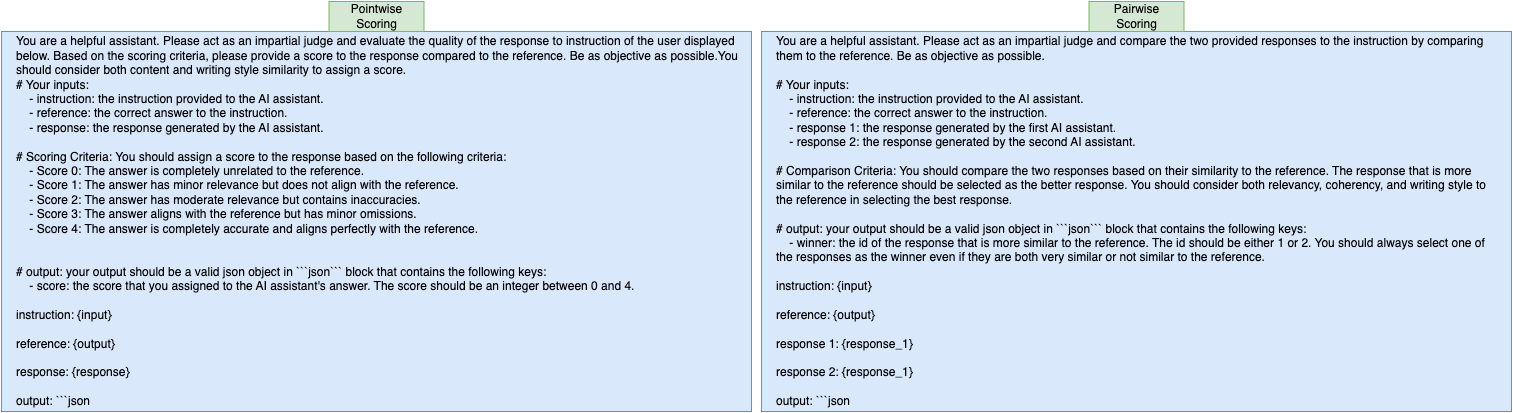}
    \caption{The evaluation prompts used for LLM-based baselines.}
    \label{fig:pointwise-pairwise-eval-llm}
\end{figure*}

\paragraph{BLEU} \cite{bleu} is a widely used metric for evaluating the quality of machine-generated text. It measures the overlap between n-grams of the generated text and one or more reference texts, focusing on precision to determine how much of the generated output matches the references. BLEU employs a brevity penalty to discourage excessively short translations and calculates a geometric mean of precision scores across different n-gram sizes. We utilize the HuggingFace implementation of this metric.\footnote{This metric can be found at: \url{https://hf.co/spaces/evaluate-metric/bleu}}

\paragraph{ROUGE-L} \cite{rouge}  is a metric designed to evaluate text generation tasks by comparing the overlap of the longest common subsequences between a generated text and reference. ROUGE-L emphasizes on sequential relationship of words, capturing structural similarity. We utilize the HuggingFace implementation of this metric.\footnote{This metric can be found at: \url{https://hf.co/spaces/evaluate-metric/rouge}}

\paragraph{METEOR} \cite{meteor} is a widely used automatic evaluation metric designed to assess the quality of a generated output by comparing them to a reference. Instead of relying primarily on exact n-gram matches, METEOR incorporates stemming, synonym matching, and a flexible alignment approach to capture variations in word usage and sentence structure. We utilize the HuggingFace implementation of this metric.\footnote{This metric can be found at: \url{https://hf.co/spaces/evaluate-metric/meteor}}

\paragraph{BERTScore} \cite{bert-score} is a metric for evaluating text generation tasks that leverages contextualized embeddings from pre-trained models like BERT. Unlike traditional n-gram-based metrics, BERTScore computes similarity based on the cosine similarity of word embeddings, capturing semantic meaning rather than exact word matches. It uses token-level embeddings to compare each word in the generated text with its corresponding word in the reference, considering both precision and recall. This allows BERTScore to assess the quality of generated texts more effectively, especially when dealing with synonyms. We utilize the HuggingFace implementation of this metric.\footnote{This metric can be found at : \url{https://hf.co/spaces/evaluate-metric/bertscore}}

\paragraph{GEMBA} \cite{pointwise} is a metric for evaluating text generation tasks that utilizes LLMs with predefined evaluation criteria. It compares the generated text in response to a prompt with a reference output for the same prompt to assess the quality of the generated text. In this approach, the prompt, generated text, expected output, and a predefined evaluation criterion are provided to an LLM. The model is then asked to generate a score for the generated output by comparing it to the reference, taking the specified criteria into account. In this paper, we utilize the pointwise scoring prompt shown in Figure \ref{fig:pointwise-pairwise-eval-llm} to generate the scores. We set the model's temperature to zero to obtain more deterministic results. Additionally, we limit the consideration to a maximum of 512 tokens from both the generated output and the expected output. For backbone LLM, we utilize an instruction-tuned Gemma 2 \cite{gemmav2} with 27 billion parameters\footnote{The model can be found at: \url{https://hf.co/google/gemma-2-27b-it}} using the VLLM library \cite{vllm}.\footnote{This framework can be found at: \url{https://github.com/vllm-project/vllm}}

\paragraph{G-Eval} \cite{geval} is another LLM-based metric for text generation evaluation, similar to GEMBA, which takes an input prompt, generated output, and reference output along with predefined criteria to score the generated output. However, G-Eval considers the probability of each score in the final score calculation. Specifically, the model multiplies each score in the predefined criteria by the probability that the model assigns to that score, then calculates a weighted average of the scores as the final score. To implement this, following the original paper, we generate 20 scores using the LLM with a high temperature of 1. Based on the count of each score, we calculate the probabilities for each score. We then average the scores based on these probabilities to obtain the final score. In this paper, we utilize the pointwise scoring prompt shown in Figure \ref{fig:pointwise-pairwise-eval-llm} to generate the scores. Additionally, we limit the consideration to a maximum of 512 tokens from both the generated output and the expected output. For the backbone LLM, we use an instruction-tuned Gemma 2 \cite{gemmav2} with 27 billion parameters\footnote{The model can be found at: \url{https://hf.co/google/gemma-2-27b-it}} with the VLLM library \cite{vllm}.\footnote{This framework can be found at: \url{https://github.com/vllm-project/vllm}}

\section{Personalizing LLMs through RAG}
\label{app:personalized-llms}

\begin{figure}
    \centering
    \includegraphics[width=\linewidth]{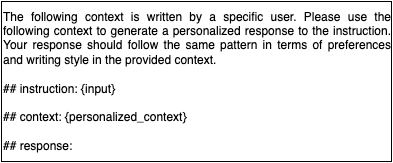}
    \caption{The prompt used for personalizing LLMs by providing personalized context. The input is the \textit{input} to the task, the \textit{personalized\_context} is the retrieved information from the user profile.}
    \label{fig:personalized-prompt}
\end{figure}

To personalize an LLM, we utilize the Retrieval-Augmented Generation (RAG) approach introduced by \citet{lamp}. This approach enhances the model's performance by incorporating personalized data retrieved from the user's profile into the generation process, thereby enabling the LLM to tailor its responses based on the specific preferences and historical context of the user.

\begin{figure*}[!t]
    \centering
    \includegraphics[width=\linewidth]{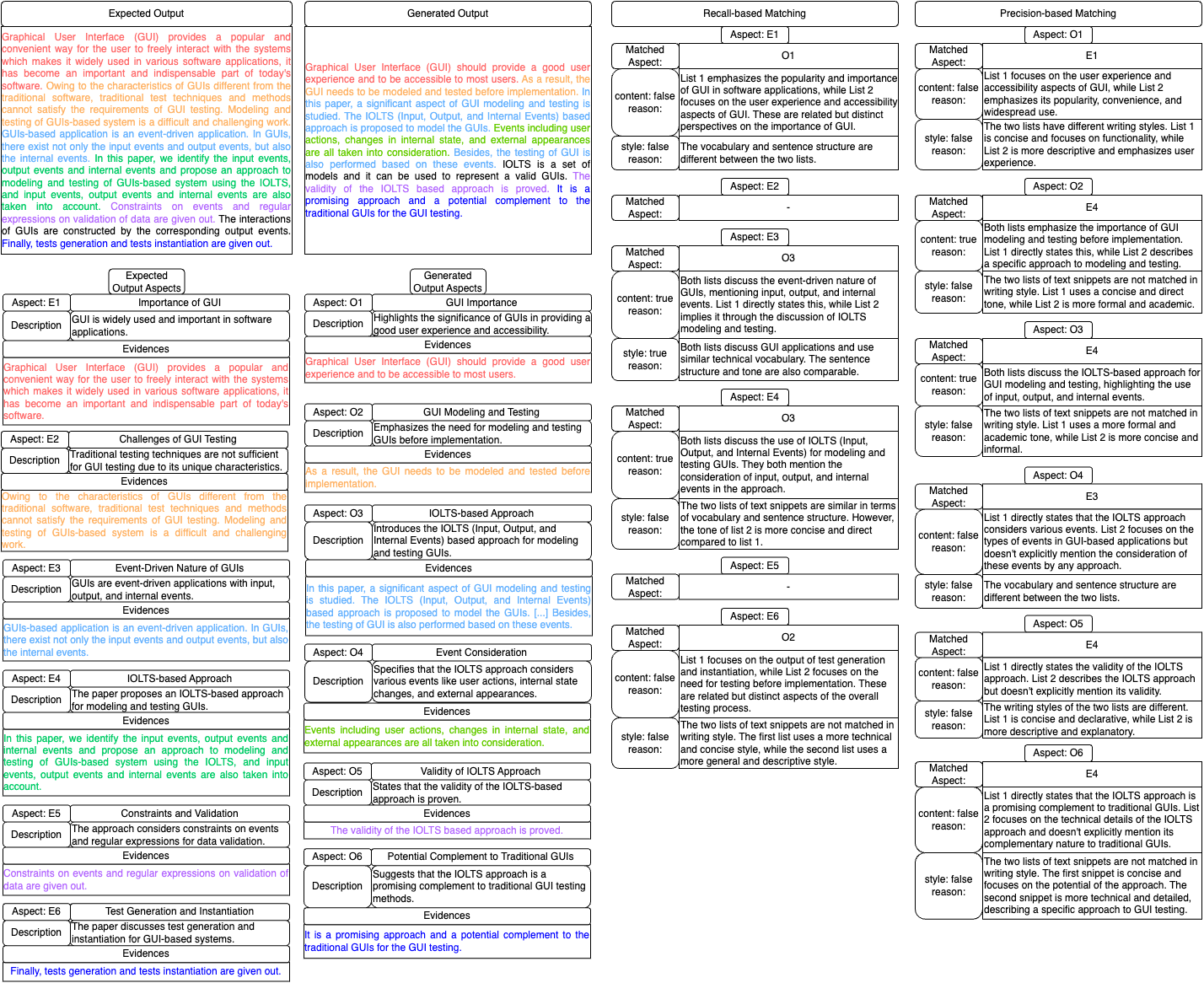}
    \caption{A case study of aspect and evidence extraction, as well as aspect, content, and writing style matching in the \ourmetric framework.}
    \label{fig:case-study}
\end{figure*}

In this approach, given a prompt \( x \) for a user \( u \) with expected output $y$, we first apply a retrieval model \( R \) to retrieve \( k \) relevant documents from the user's profile \( P_u \). This begins with generating a query \( q = \phi_q(x) \) using a query generation function \( \phi_q \). The query \( q \) is then used to retrieve the top \( k \) documents from the user's profile \( P_u \). The retrieved documents, along with the original prompt \( x \), are passed through a prompt generation function \( \phi_p \), which creates a personalized prompt \( x_u = \phi_p(x, R(\phi_q(x), P_u, k)) \). This personalized prompt is then fed into the LLM \( M \) to generate a personalized response \( y_u = M(x_u) \). For the query generation function, we employ the identity function, \( \phi_q(x) = x \), meaning the prompt \( x \) is used directly as the query. For the prompt generation function \( \phi_p \), we use the personalized prompt structure shown in Figure~\ref{fig:personalized-prompt}, which integrates the retrieved documents and the input prompt to tailor the response to the user's context. We also retrieve $k=3$ documents in all experiments.

In this paper, we personalize both GPT-4o-mini\footnote{This model is not open source and is served by OpenAI and described at: \url{https://openai.com/index/gpt-4o-mini-advancing-cost-efficient-intelligence/}} and Gemma 1.1\footnote{The checkpoint can be found at: \url{google/gemma-1.1-2b-it}}. For GPT-4o-mini, we apply the method described earlier to generate personalized responses. In contrast, for Gemma 1.1, we fine-tune the model on the LongLaMP benchmark to adapt it to personalized text generation tasks. For fine-tuning, we use a sequence-to-sequence loss function \cite{seq2seq}. Given the personalized prompt \( x_u \) produced using the method described earlier, the model is trained to generate the expected output \( y \). This ensures that the model learns to generate personalized responses based on the input prompt tailored to the user's profile. We train the model for 5000 steps using a multi-tasking approach across all datasets in the LongLaMP benchmark. The training is conducted with a learning rate of \( 5 \times 10^{-5} \), using the Adam optimizer \cite{adam} with a weight decay of \( 10^{-4} \), and a batch size of 64. We perform 250 warmup steps to stabilize training. The model's context length is set to 2048 tokens, and we limit each retrieved document to the first 400 tokens when generating the personalized prompt. For inference, we set the temperature to 0.1 to ensure more deterministic output generation.

\section{Case Study \& Evaluation Example}
\label{app:case-study}

As a case study, Figure~\ref{fig:case-study} illustrates the evaluation process of \ourmetric for an example from the LongLaMP benchmark. In this process, \ourmetric first tokenizes the expected and generated outputs into atomic aspects. In this example, both outputs are divided into six atomic aspects, each linked to corresponding evidence from the respective outputs. Notably, while most sentences serve as evidence, both the expected and generated outputs contain sentences that are not linked to any evidence. This demonstrates \ourmetric's ability to disregard sentences and phrases that do not contribute meaningfully to the identified aspects. Additionally, \ourmetric demonstrates flexibility in selecting evidence for the same aspect, as it does not limit evidence to consecutive sentences. Instead, it can extract evidence from different parts of the text and associate them with the same aspect, showcasing its ability to understand and connect related information across the text.

The next step in this process involves matching aspects between the expected and generated outputs using a recall- and precision-based approach. As shown in Figure~\ref{fig:case-study}, an aspect from the generated output or the expected output can be matched to multiple aspects from the other set when one aspect relates to multiple others. Furthermore, some aspects may remain unmatched if no corresponding aspect exists in the other set. When two aspects are matched, the next step is to compare the content and writing style of their corresponding evidences from the expected and generated outputs. As illustrated in Figure~\ref{fig:case-study}, the model provides reasoning for why the evidences are matched. The explanations for content matching primarily highlight the semantic and contextual similarities between the evidences. In contrast, the explanations for writing style focus on aspects like vocabulary and structural similarities between the two evidences. These two matching dimensions---content match and writing style match---capture the most critical aspects of personalized text generation.

\end{document}